# Drift-Adjusted And Arbitrated Ensemble Framework For Time Series Forecasting

Anirban Chatterjee, Subhadip Paul, Uddipto Dutta, Smaranya Dey
WalmartLabs
Bangalore, India
Email:{Anirban.Chatterjee, Subhadip.Paul0, Uddipto.Dutta}@walmartlabs.com, Smaranya.Dey@walmart.com

*Abstract*—Time Series Forecasting is at the core of many practical applications such as sales forecasting for business, rainfall forecasting for agriculture and many others. Though this problem has been extensively studied for years, it is still considered a challenging problem due to complex and evolving nature of time series data. Typical methods proposed for time series forecasting modeled linear or non-linear dependencies between data observations. However it is a generally accepted notion that no one method is universally effective for all kinds of time series data Attempts have been made to use dynamic and weighted combination of heterogeneous and independent forecasting models and it has been found to be a promising direction to tackle this problem. This method is based on the assumption that different forecasters have different specialization and varying performance for different distribution of data and weights are dynamically assigned to multiple forecasters accordingly. However in many practical time series data-set, the distribution of data slowly evolves with time. We propose to employ a re-weighting based method to adjust the assigned weights to various forecasters in order to account for such distribution-drift. An exhaustive testing was performed against both real-world and synthesized time-series. Experimental results show the competitiveness of the method in comparison to state-of-the-art approaches for combining forecasters and handling drift.

## I. INTRODUCTION

Time series forecasting is one of the most challenging yet interesting tasks in time series analysis due to the evolving structure of time series data, which may involve non-stationary and complex process [2], [3], [6]. Over the period of time, forecasting has attracted attention not only of academia but also of industrial communities due to several reasons. It has always been at the core of many real-time decision-making and planning processes across various applications such as financial planning, workforce planning, demand forecasts, traffic forecasts and many others [3].

Several Machine Learning methods have been proposed for time series forecasting. One class of proposed methods treat time series data as an ordered sequence for forecasting while another class formulates the forecasting task as a regression task by creating an embedding out of the time series data. However, it is a generally accepted notion that no one method is equally effective for all kinds of time series data. It acted as one of the primary motivations behind the exploration of the option to construct an ensemble of multiple heterogeneous time series forecasting models in dynamic fashion. The process to construct an ensemble is broadly split into three parts [5] - (i) build base models to forecast future observation given the observation data in the recent past (ii) build meta learners which will predict the expected error from the forecasters at various time instants (iii) assign weights dynamically to different forecasters on the basis of the prediction of meta-learners.

Most of the existing strategies of constructing ensemble focus on various ways to assign weights to forecasters. Ensemble strategies can be grouped into three main families [5]. The first family relies on voting approaches using majority or weighted average votes to decide the final output. The second main family of methods propose a cascading strategy, where outputs of the base models are iteratively included once at a time, as new features in the training set. The third group introduced stacking paradigm [10] where a meta-learning approach is employed to combine the available forecasts. This method implicitly learns an adequate combination by modelling inter-dependencies between models.

Another important aspect of learning ensemble for time series data is to be able to address time-evolving nature of data. This can be achieved by building dynamic ensemble frameworks through adaptive assignment of weights to different forecasters which can account for distribution-drift of data over the period of time.

Our proposed approach is similar to [6] in spirit but with an enhancement to deal with evolving nature of data. In a nutshell, our method segments the time-series data for training $T$ into a number of non-overlapping segments, $T_1, T_2, ...T_m$ and extracts a $k$ dimensional embedding from each and every segment. To execute the task to forecast the value at time-instant $t + 1$ when the data is available upto time-instant $t$, it checks how the forecasters performed in different time segments, $T_i$. Normalized weights are assigned to different forecasters according to the similarity of the data-distribution of recent past with that of various training time-segments, $T_i$, and the performances of forecasters in those training time-segments. This approach is effective whenever the time-series data exhibit a consistent distribution over the period of time. In case there is a drift in distribution, we need a way to estimate the performance of various forecasters with respect to the changed pattern of data from their performances with respect to slightly different pattern in the past. This estimation of performance is used in assignment of weight to the forecasters. We propose to employ a re-weighting based approach to maximize the similarity between the distribution of two time-segments and adjust the assigned weights accordingly.

In summary the contributions of this papers are:
- Our proposed approach accounted for gradual drift in data distribution while assigning weights to forecasters in ensemble framework.
- A thorough analysis to understand when such approach may not work optimally and proposal of a way to mitigate that issue.

We start by outlining the related work in Section II; some preliminary concepts in section III; the methodology is addressed in Section IV, where we described the proposed algorithm along with the analysis of its various components; data used to carry out the experiment have been described in section IV; experimental results and corresponding analysis are presented and discussed in Section VI. Finally, Section VII concludes the paper.

## II. RELATED WORKS

In this section we briefly discuss the state-of-the-art methods for adaptively constructing ensemble of models for time series forecasting along with their characteristics and limitations and highlight our contributions.

The simple average of the predictions of the available forecasting models have been shown to be a robust combination method [6], [7], [8]. Nonetheless, other more sophisticated approaches have been proposed.

### A. Windowing Strategy for Model Combination

AEC is a method for adaptively combining forecasters [12]. It uses a weighting strategy to combine forecasters according to their past performance. It includes a forgetting factor to give more importance to recent values. According to [13], the stock returns model have short-lived period of predictability. Therefore it proposes an adaptive combination based on the recent $R^2$ values of the forecasters. If all the models exhibited poorly on the $R^2$ values in the recent past, forecast value of the combined model tends to be closer to the mean of those observations. Otherwise, combined model outputs the arithmetic mean of the prediction values of all the base-learners.

Newbold and Granger proposed a method of linearly weighting models on the basis of their performance in the recent past [14]. Even though the proposed heuristics to combine the forecasters is adaptive, it is incremental and dependent of sliding summary statistics in the recent past. Conversely, we not only explored the entire data space to compute the expertise of the base-forecasters to forecast the value of the next time-instant but also accounted for gradual evolution of data distribution over the period of time.

### B. Meta-learning Strategies for Model Combination

Meta-learning is a way to learn a model about the learning process itself [9] Several meta-learning based approaches have been proposed to improve the construction of the ensemble of forecasters [10], [11]. A meta-learning strategy, called arbitration, has been proposed in [6], which dynamically assigns weights to various forecasting models according to their expertise given the input data. A set of meta-learners, one fore every base forecasting model, is employed to model the loss of its base counterpart. It essentially relies on weighting all the forecasters rather selecting a few of them. It addresses several drawbacks of earlier ensemble frameworks for forecasting, such as inefficient use of the available data, by using OOB samples from the training set. It introduces more robust combination rule by using a committee of recent well performing models and weighing them using a softmax function.

Even though it addressed many important aspects of construction of ensemble of time series forecasting models, its effectiveness comes under severe limitation when the distribution of time series data slowly evolves with time. It becomes difficult to compute the expertise of forecasters and assign weights accordingly while forecasting the value at next time-instant, if the distribution of the time series data in the recent past is substantially different from the data observed in other parts of data space. We explored re-sampling based adjustment of assigned weights to forecasters in case the divergence between the distribution of data in the recent past and other portion of the data space which is most similar to the one in recent past, exceeds a limit.

## III. PRELIMINARY CONCEPTS

### A. T-Test

If there is a sample with sample mean and standard deviation are $\hat{\mu}$ and $\hat{\sigma}$ respectively and population mean is $\mu$, then the random variable, represented by $\sqrt{n-1}\frac{\hat{\mu}-\mu}{\hat{\sigma}}$ follows t-distribution with $n-1$ degrees of freedom.

Let us consider a *t-statistic*,

$$T = \sqrt{n-1}\frac{\hat{\mu}-\mu_0}{\hat{\sigma}} \quad (1)$$

The statistic will behave differently if true unknown mean $\mu = \mu_0$, $\mu < \mu_0$ or $\mu > \mu_0$. If $\mu = \mu_0$, then $T \sim t_{n-1}$. If $\mu < \mu_0$, we can write,

$$T = \sqrt{n-1}\frac{\hat{\mu}-\mu}{\hat{\sigma}} + \sqrt{n-1}\frac{\mu-\mu_0}{\hat{\sigma}} \to -\infty \quad (2)$$

Notice that the first term in above expression follows t-distribution. Similarly, $\mu > \mu_0$ results in $T \to +\infty$. Therefore, we can make decision about our hypothesis based on the values of $T$.

I.($H_0 : \mu = \mu_0$) In this case, the indication that $|T|$ is too large results in rejection of hypothesis $H_0$. Therefore, we consider the decision rule,

$$\delta = \begin{cases} H_0, & \text{if } -C < T < C. \\ H_1, & \text{if } |T| > C. \end{cases} \quad (3)$$

The value of $C$ depends on the significance level $\alpha$. We would like to have,

$$\mathbb{P}(\delta = H_1|H_0) = \mathbb{P}(|T| > C|H_0) \leq \alpha \quad (4)$$

*B. Paired T-Test*

Paired T-Test is carried out to test if the mean of the populations of two samples $X$ and $Y$ are same. We can construct a random variable $Z_i = X_i - Y_i$ assuming $X_i$ and $Y_i$ are i.i.d. If we conduct a simple t-test on $Z_i$ to check if $\mu_z = 0$, it becomes equivalent to check if the mean of the population of $X$ and $Y$ are same.

## IV. METHODOLOGY

The idea of combining multiple forecasters is described in Algorithm1.

The algorithm has three major components:

*A. Learning Base-Level Models*

In the offline learning phase we train $n$ individual forecasters independently. The objective of each and every model $M_i, i \in 1, ..., n$ is to predict the next value $y_{t+1}$ when the data upto time instant $t$ is available. It is expected that the heterogeneity in models will result in different inductive bias and different expertise against different pattern of time-series data.

*B. Learning Meta-Level Models*

In meta-learning step, the expertise of each forecaster across the data space is modelled. In other words, the meta-level models predict the error that different forecasters might admit while forecasting the point $y_{t+1}$ at time-instant $t+1$ when data is available upto time-instant $t$.

*C. Drift Adjustment*

It is the primary contribution of this paper. In general sense, in the context of time series, drift essentially means divergence between the probability distribution, $Pr$ and $Pr'$ of two time segments. However, for the sake of simplicity, we assumed the data in two different segments of time series exhibit similar kind of distribution but with a different mean. As we have finite samples from both the distributions, we will naively replace mean by empirical sample average.

We may formulate the following quadratic problem to minimize the drift in mean between the data samples $x^{T_i}$ and $x^{T_j}$ of two time segments, $T_i$ and $T_j$, such that $x^{T_i} \sim Pr$ $x^{T_j} \sim Pr'$,

$$\min_w ||\frac{1}{n}\sum_{k=1}^n w_k x_k^{T_i} - \frac{1}{n}\sum_{k=1}^n x_k^{T_j}||^2$$
$$\text{s.t.} \quad w_i \in [0, B], \quad |\frac{1}{n}\sum_{i=1}^n w_i - 1| \leq \epsilon \quad (5)$$

The former constraint helps in regularization while the later ensures $w(x)Pr(x)$ is close to the target probability distribution with respect to its mean.

Observe that a more generalized and continuous version of the above quadratic problem will be,

$$\min_w ||E_{x \sim Pr}(w(x) \cdot x) - E_{x \sim Pr'}(x)||^2$$
$$\text{s.t.} \quad w(x) > 0, \quad E_{x \sim Pr}(w(x)) = 1 \quad (6)$$

**Comment 1.** Problem 2 will yield $w(x)Pr(x) = Pr'(x)$

Notice that both the objective function and constraints are convex in Problem 2. For the time being if we ignore the constraint $E_{x \sim Pr}(w(x)) = 1$, the objective function minimizes when

$$(E_{x \sim Pr}(w(x) \cdot x) - E_{x \sim Pr'}(x))(\int xPr(x)dx) \\ = 0 \quad (7)$$

It essentially implies

$$\int xw(x)Pr(x)dx = \int xPr'(x)dx \quad (8)$$

Constraints, $E_{x \sim Pr}(w(x)) = 1$ and $w(x) > 0$, imply that $\int w(x)Pr(x)dx = 1$. It means $w(x)Pr(x)$ is a proper probability distribution function as it sums upto 1 as well as takes $\geq 0$ value $\forall x$. Therefore, (4) implies $E_{x \sim w(x)Pr(x)}(x) = E_{x \sim Pr'(x)}(x)$ i.e. $w(x)Pr(x)$ and $Pr'(x)$ are distributions with same mean. It is indeed our final objective that is re-weighting the data points of one time-segment such that the distribution of transformed data points has same mean as that of the other.

**Comment 2.** If $w(x) \in [0, B]$, then the empirical mean $\frac{1}{m}\sum_{i=1}^m w(x_i)$ converges to a normal distribution with mean $\int w(x)Pr(x)dx$ and its standard deviation is bounded by $\frac{B}{2\sqrt{m}}$.

It is a direct result from central limit theorem and Popoviciuś inequality.

**Comment 3.** If $||x|| \leq R \; \forall x$, the the following inequality holds with probability $1 - \delta$,

$$|\frac{1}{n}\sum_{k=1}^n w_k x_k^{T_i} - \frac{1}{n}\sum_{k=1}^n x_k^{T_j}| \leq (1 + \sqrt{\log(2/\delta)})R\sqrt{\frac{(B^2+1)}{n}} \quad (9)$$

Refer to [16] for the proof of observation 3. What it implies is, unless and until there is high number of samples, there is no guarantee that after convergence, the mean of time-segments will be very similar.

**Comment 4.** The following inequality holds with probability $1 - \delta$,

$$|\frac{1}{n}\sum_{k=1}^n w_k - 1| \leq B\sqrt{\log(2/\delta)/2n} \quad (10)$$

Above is a direct result from "Hoeffding's Inequality". Notice that when the random variable $w$ is continuous as in (2), one of the constraints is set as $E_{x \sim Pr}(w(x)) = 1$. However in the empirical optimization the absolute difference between empirical mean of $w$ and its expected value, i.e. 1, has an upper-bound which is directly proportional to $B$ and $\frac{1}{\sqrt{n}}$. High value of $B$ due to high standard deviation of $w$ may cause the empirical mean of optimal choice of $w$ to be far from 1 and in that case $w(x)Pr(x)$ won't be close to $Pr'(x)$. That situation may arise when the two time-segments have large difference in their mean value.

However by tuning the hyper-parameters such as the upper-bound of the sample-weights and collecting as many samples in multiple time-segments as possible, it is possible to match the mean of two time-segments with reasonable accuracy by re-weighting the samples of one of the segments.

## V. Dataset

### A. Order Data from offline retail store

Daily aggregated order data, as visualized in Fig 1, was collected from an offline retail store. A constant number was added to every data-point in the time-series to maintain its confidentiality. As same number has been added to every data point in time-series, it does not alter the pattern of the data.

### B. Synthetic Data

Time-series data has been synthesized such a way that the mean, between multiple segments in both the test and training data, differs. Refer to Fig 2 for the visualization of the same.

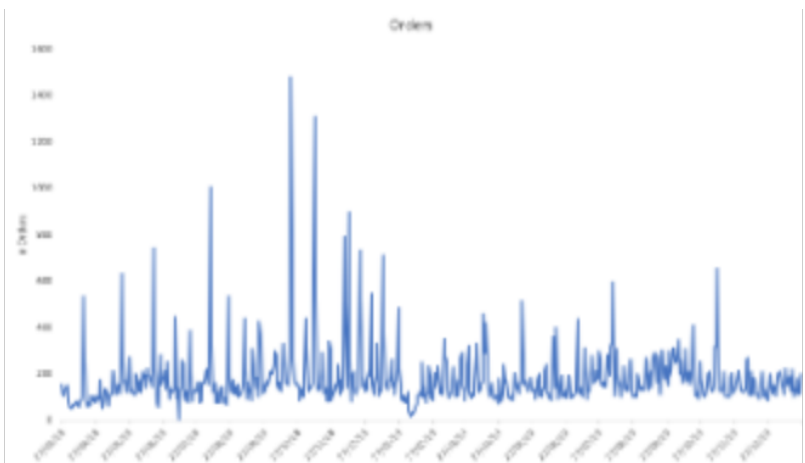

Fig. 1. Masked Order Data from retail store

## VI. Results and Analysis

In the proposed ensemble framework three different kinds of forecasters, ARIMA, BSTS (Bayesian Structural Time-Series) and Neural-Network based forecasting model, have been used. State-of-the-art argorithm in [6] to combine forecasters has been considered as the benchmark to evaluate the result.

**Algorithm 1:** Ensemble Forecasting Algorithm

Input;
$\{X_t\}$ - Time-series data, $h$ - Forecast-period;

Partition $X_t$ into $n$ non-overlapping slices, $X_{t_1,t_1+L}, ..., X_{t_n,t_n+L}$, of size $L$ from the end;

**for** i in 2:n-1 **do**
    **if** *p-value of paired t-test between* $X_{t_i,t_i+L}, X_{t_n,t_n+L} \geq 0.5$ **then**
        // $q_{x/k}$ is $x^{th}$ quantile of $X_{t_i,t_i+L}$
        Generate $k+1$ *features from* $X_{t_i,t_i+L}$ *as :* $\{q_0, q_{1/k}, ..., q_{k-1/k}, q_1\}$
    **else**
        Find $\alpha$ to minimize
        $\frac{1}{m}||\sum_{j=1}^{m} \alpha_j X_{t_i,t_i+L}(j) - \sum_{j=1}^{m} X_{t_n,t_n+L}(j)||^2$
        *such that* $\alpha_j > 0 \;\; \forall j$ *and* $\frac{1}{m}\sum_{j=1}^{m} \alpha_j = 1$
    **end**
**end**

Train model $M_k$ on original dataset to get one-point forecast // $M_k \in \{ARIMA, BSTS, Neural-net\}$

**for** j in 2:n **do**
    Train each model $M_k$ on $\{\alpha_t X_{t_j, t_j+L}\}$ to find one-point forecast. ( $\alpha_t = 1 \; \forall t$ if at the $j^{th}$ slice, drift adjustment didn't occur)

    Calculate the $MAPE$ for model $M_k$, say $e_k$, for one-point forecast.
**end**

For each model $M_k$, fit a regression tree with the MAPEs of each slice as dependent variable and the features generated for each slice as the regressors to predict the MAPE of $M_k$, $\tilde{e}_k$, at the newest slice of data.

Weight assigned to the model $M_k$ in ensemble = $\frac{\exp(-\tilde{e}_k)}{\sum_i \exp(-\tilde{e}_k)}$

Fig 3 and Fig 4 visualize the forecasting performance of two ensemble models against the part of order data and synthesized data, held-out for testing The shaded region of fig 3 and fig 4 suggests that the ensemble model with drift-adjustment offers improvement in accuracy over ensemble model without drift-adjustment However the degree of improvement is observed to be substantial for order data but marginal for synthetic data.

A reasonable explanation of marginal improvement for synthetic data can be elicited from Comment 3 as mentioned above. Notice that for two time-segments, $T_i$ and $T_j$, with mean $\mu_1$ and $\mu_2$ respectively such that $\mu_1 \sim \mu_2$, $w(x) \longrightarrow 1 \forall x$ which results in $B \longrightarrow 1$. Comparatively lower value of $B$ pulls down the upper bound of $|\frac{1}{n}\sum_{k=1}^{n} w_k x_k^{T_i} - \frac{1}{n}\sum_{k=1}^{n} w_k x_k^{T_j}|$, close to $\frac{\sqrt{2}R}{\sqrt{n}}(1+\sqrt{\log(2/\delta)})$. It can be pulled down further by increasing the number of samples.

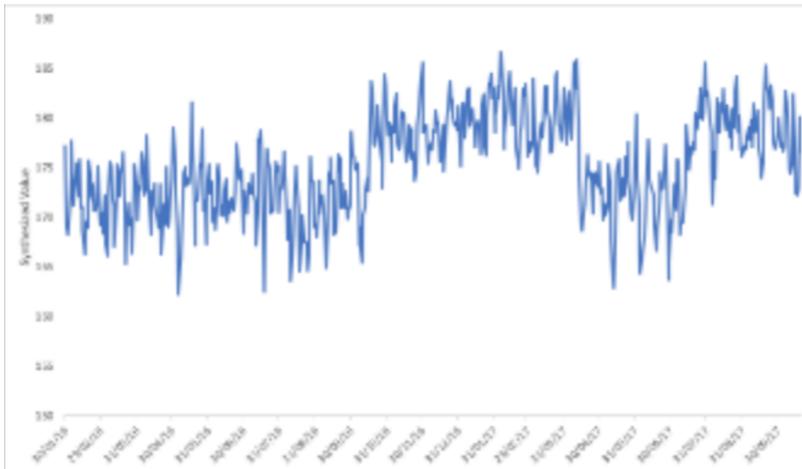

Fig. 2. Synthesized time-series data

On the other hand, $|\mu_1 - \mu_2| >> \epsilon$ leads to higher standard-deviation of $w$ which essentially means $w_{max} >> 1$ thereby leading to $B >> 1$ as $w \in [0, B]$. Higher value of $B$ pushes up the upper bound of $|\frac{1}{n}\sum_{k=1}^{n} w_k x_k^{T_i} - \frac{1}{n}\sum_{k=1}^{n} w_k x_k^{T_j}|$. Therefore, even after the optimization algorithm of Problem 1, as mentioned above, converges finding the optimal $w$, the point of convergence of the objective function can be far from 0 with high probability. The only way to mitigate this, is to increase the number of samples, $m$, in both the time-segments.

In the context of synthetic data and order data, it is clearly visible from Fig 3 and Fig 1 that the mean of multiple segments of synthesised data have much more difference

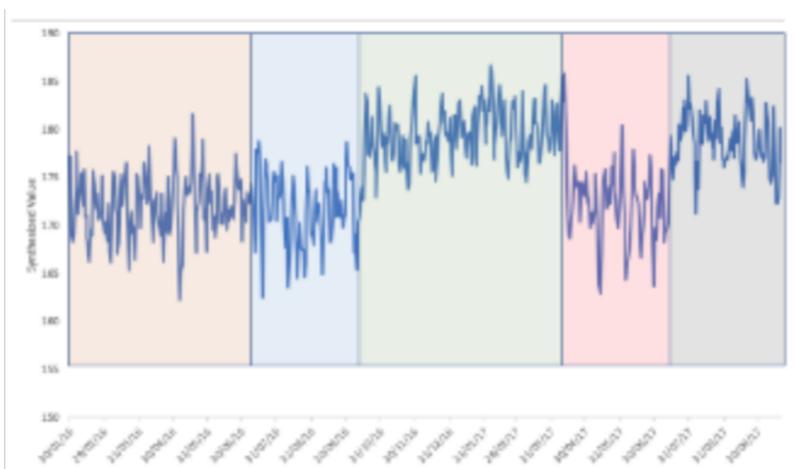

Fig. 3. Heterogeneous segments of synthesized time-series data

than order data. This issue can be mitigated by synthesizing more samples in all the time segments.

## VII. CONCLUSION

We presented an adaptive ensemble method with drift-adjustment for time-series forecasting tasks. Our strategy to account for gradual evolution of distribution of time-series data, provides a way to deal time-series with local stationarity. Therefore, the proposed model is adaptive to time-variability of the underlying process which generates data in the time-series.

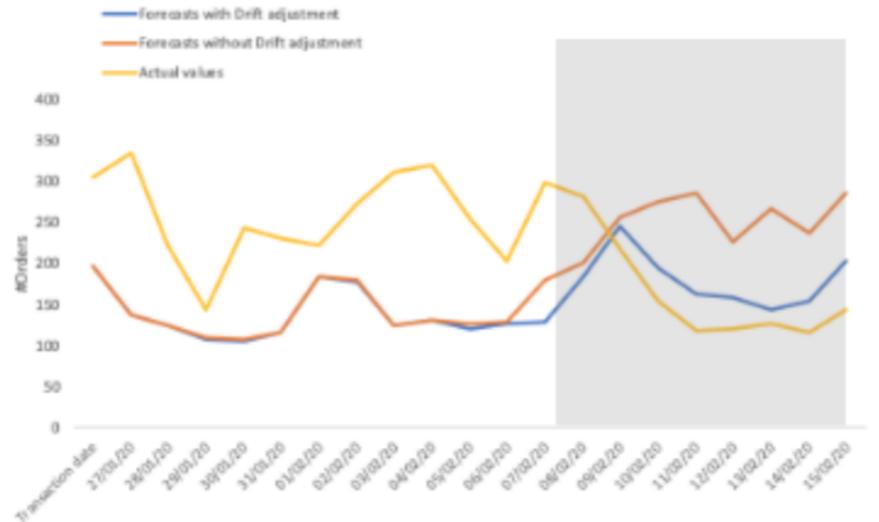

Fig. 4. Forecasting in test-set of order data

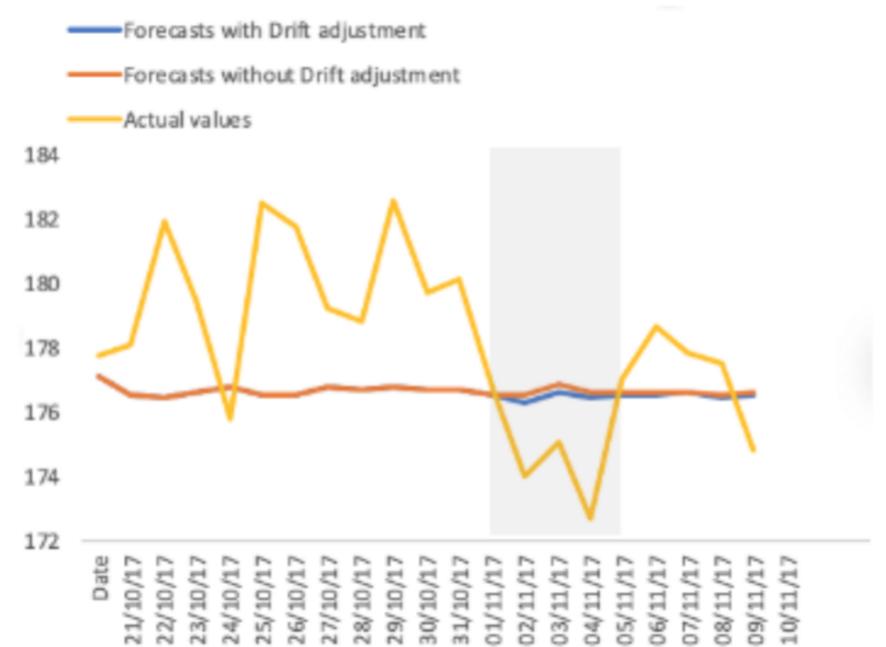

Fig. 5. Forecasting in test-set of synthesized data

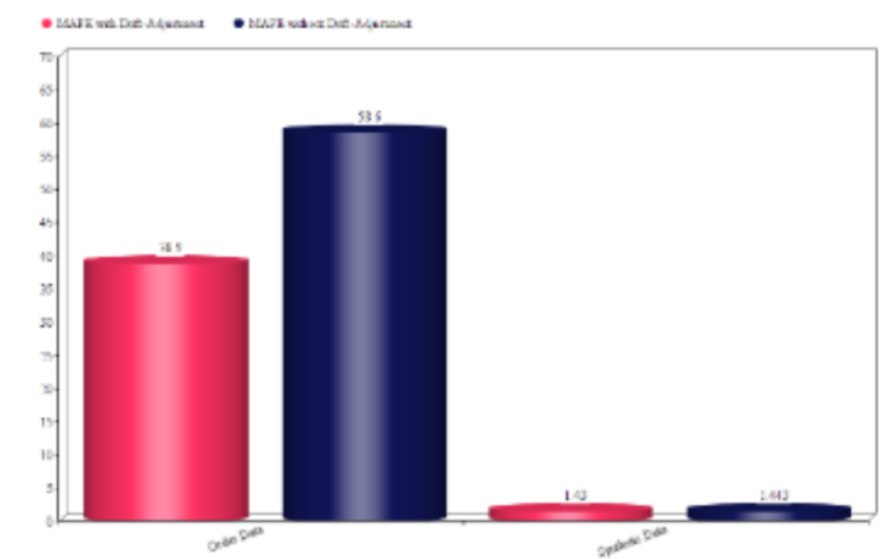

Fig. 6. Forecasting accuracy with and without drift-adjustment

The proposed enhancement to ADE [6] is motivated by this observation that in many practical applications, such as sales data of an organization in emerging market, time-series data undergoes gradual but permanent change. Therefore along with learning the specialization of multiple forecasters, it is equally important to be able to estimate the performance of forecasters when data-pattern is evolving. We proved the competitiveness of our approach in both real-world time-series data along with synthesized time-series data. We also analyzed why the improvement in accuracy from drift-adjusted forecasting model may not be up to the mark when there is an abrupt change in data-pattern in the time-series and suggested a way to mitigate it. We plan to work out a more robust and explainable solution to adapt the proposed method to a time-series exhibiting abrupt change in data pattern.

## Acknowledgment

We gratefully acknowledge the support from WalmartLabs, Bangalore to continue this work. We also thank Gayatri Pal and Vijay Agneeswaran for their valuable inputs and suggestions at times.